\pdfoutput=1

\documentclass[11pt]{article}

\usepackage{EMNLP2022}

\usepackage{times}
\usepackage{latexsym}

\usepackage[T1]{fontenc}

\usepackage[utf8]{inputenc}

\usepackage{microtype}

\usepackage{inconsolata}

\usepackage{booktabs}       
\usepackage{amsfonts}       
\usepackage{nicefrac}       
\usepackage{microtype}      
\usepackage{xcolor}         
\usepackage{url}
\usepackage{amsmath}
\usepackage{graphicx} 
\usepackage{multirow}
\usepackage{multicol}
\usepackage{algorithm}    
\usepackage{algpseudocode}
\usepackage{amssymb}
\usepackage{ucs}
\usepackage{enumitem}
\usepackage{color}
\usepackage{subcaption}

\title{EdgeFormer: A Parameter-Efficient Transformer for \\ On-Device Seq2seq Generation}

\author{Tao Ge ~~~~~~~~Si-Qing Chen ~~~~~~~~Furu Wei \\
Microsoft \\
{\tt \{tage,sqchen,fuwei\}@microsoft.com}
}

\begin{document}

\maketitle

\begin{abstract} 
We introduce \textsc{EdgeFormer} -- a parameter-efficient Transformer for on-device seq2seq generation under the strict computation and memory constraints. Compared with the previous parameter-efficient Transformers, \textsc{EdgeFormer} applies two novel principles for cost-effective parameterization, allowing it to perform better given the same parameter budget; moreover, \textsc{EdgeFormer} is further enhanced by layer adaptation innovation that is proposed for improving the network with shared layers.
Extensive experiments show \textsc{EdgeFormer} can effectively outperform previous parameter-efficient Transformer baselines and achieve competitive results under both the computation and memory constraints. Given the promising results, we release \textsc{EdgeLM}\footnote{\url{https://github.com/microsoft/unilm/tree/master/edgelm}} -- the pretrained version of \textsc{EdgeFormer}, which is the first publicly available pretrained on-device seq2seq model that can be easily fine-tuned for seq2seq tasks with strong results, facilitating on-device seq2seq generation in practice.

\end{abstract}

\section{Introduction}\label{sec:intro}
On-device modeling draws increasing attention for its unique advantages~\citep{dhar2019device}. On the other hand, strict resource constraints prevent many neural networks performing well in the on-device setting. In Natural Language Processing (NLP), on-device sequence-to-sequence (seq2seq) generation remains challenging, especially for the Transformer~\citep{vaswani2017attention} under strict resource constraints in both computation and memory.

\begin{figure}[t!]
    \centering
    \includegraphics[width=8cm]{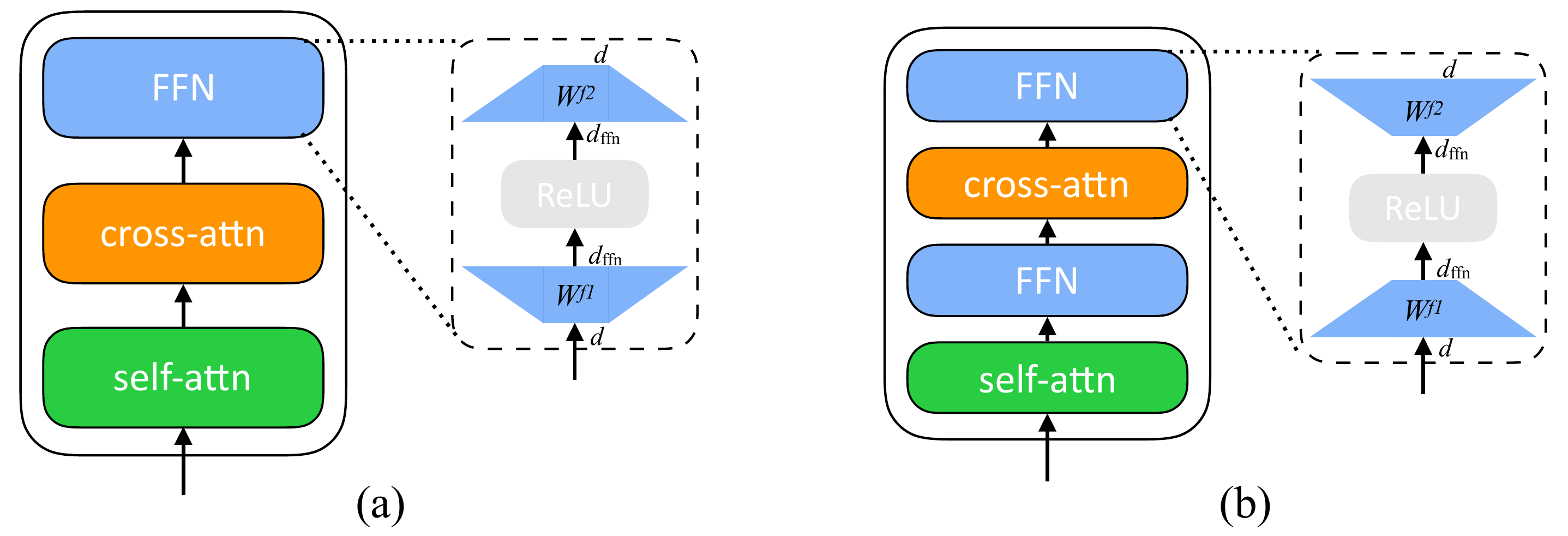}
    \caption{\textbf{(a)} Vanilla Transformer decoder layer in which $d_{\textrm{ffn}}>d$; \textbf{(b)} Interleaved Transformer decoder layer with shared lightweight FFNs in which $d_{\textrm{ffn}}<d$.}
    \label{fig:interleaved}
\end{figure}

To customize the Transformer for seq2seq tasks in the on-device setting, we propose \textsc{EdgeFormer} -- a novel parameter-efficient Transformer of the encoder-decoder architecture. \textsc{EdgeFormer} is structurally similar to the standard Transformer with a deep encoder and shallow decoder, but with a modification that it uses an interleaved decoder with shared lightweight feed-forward networks, as shown in Figure \ref{fig:interleaved}. The modified decoder architecture allows \textsc{EdgeFormer} to apply two novel principles that we propose for cost-effective parameterization: 1) encoder-favored parameterization that suggests we parameterize the encoder using as many parameters as possible; 2) load-balanced parameterization that suggests we balance the load of model parameters to avoid them being either underused or overused in a neural network with shared parameterization.

\begin{figure*}[t]
    \centering
    \includegraphics[width=14cm]{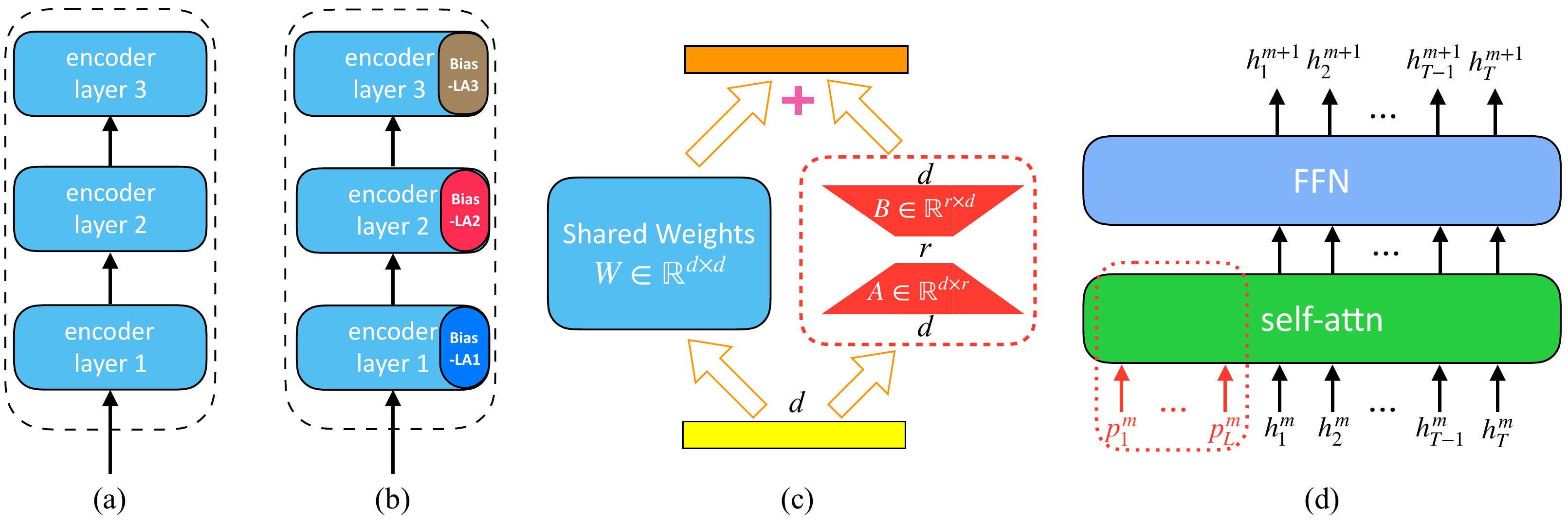}
    \caption{\textbf{(a)} Encoder layers with shared weights (the same color) without layer adaptation: the tied weights undermine the specialities of encoder layers to process their specific inputs; \textbf{(b)} Bias-based Layer Adaptation (Bias-LA) employs free bias terms to adapt layers with tied weights to fit their specific roles well; \textbf{(c)} Adapter-LA uses a layer-specific LoRA adaptation block with rank $r<d$ for layer adaptation; \textbf{(d)} Prefix-LA uses $L$ layer-specific tokens (i.e., learnable parameters) as the prefix (dotted square) to adapt the $m$th layer.}
    \label{fig:full_LA}
\end{figure*}

In addition to cost-effective parameterization, \textsc{EdgeFormer} proposes and applies layer adaptation to further improve the model with tied layers, as Figure \ref{fig:full_LA} shows. Inspired by parameter-efficient task transfer, we investigate 3 efficient layer adaptation approaches for improving the performance with negligible cost. We show \textsc{EdgeFormer} (with fewer than 10 million model parameters) largely outperforms the strong 
\textsc{Universal Transformer} baselines in the on-device setting with competitive results, and the int8-quantized \textsc{EdgeFormer} can perform high-quality on-device seq2seq generation within around 100ms latency (20-30 sequence length on average) using two mid-to-high-end CPU cores and less than 50MB RAM.

The contributions of this work are three-fold:
\begin{itemize}
    \item This paper is one of the earliest work that formally studies on-device seq2seq generation by discussing its challenges and defining a practical setting with appropriate resource constraints for the evaluation.
    \item We propose \textsc{EdgeFormer}, a parameter-efficient Transformer with novel cost-effective parameterization and layer adaptation, achieving the state-of-the-art result in the on-device seq2seq generation setting under strict computing and memory resource constraints.
    \item We introduce and release \textsc{EdgeLM} (the pretrained \textsc{EdgeFormer}) -- the first publicly available pretrained on-device seq2seq model that can be easily fine-tuned for seq2seq tasks with strong results, which can largely reduce the effort for delivering a powerful on-device seq2seq model in practice.
\end{itemize}

\section{Background: Transformer}\label{sec:background}
\subsection{Architecture}
The Transformer follows the encoder-decoder architecture. The Transformer encoder consists of a stack of encoder layers, each of which has a self-attention module parameterized by projection matrices for the query, key, value and output: [$W^Q$, $W^K$, $W^V$, $W^O$] whose shapes are all $d 
\times d$, followed\footnote{For simplicity, we omit discussing the layer normalization and residual connection that are not related with this work.} by a feed-forward network (FFN) parameterized by $W^{f1} \in \mathbb{R}^{d \times d_{ffn}}$ and $W^{f2} \in \mathbb{R}^{d_{ffn} \times d}$.

The Transformer decoder consists of a stack of decoder layers whose architecture is similar to an encoder layer except for an additional cross attention module between self-attention and FFN. 

In summary, we understand that the main parameters in an encoder layer $i$ are:
\begin{displaymath}
\small
\boldsymbol{\Phi}_{e_i}=[W^{\{Q,K,V,O\}}_{e_i}, W^{f1}_{e_i}, W^{f2}_{e_i}]
\end{displaymath}
$|\boldsymbol{\Phi}_{e_i}|=4d^2+2d\times d_{\textrm{encffn}}$. For a decoder layer $j$, its main parameters are:
\begin{displaymath}
\small
\boldsymbol{\Phi}_{d_j}=[W^{\{Q,K,V,O\}}_{d_j}, W^{\{\mathcal{Q},\mathcal{K},\mathcal{V},\mathcal{O}\}}_{d_j}, W^{f1}_{d_j}, W^{f2}_{d_j}]
\end{displaymath}
where $W^{\{\mathcal{Q},\mathcal{K},\mathcal{V},\mathcal{O}\}}_{d_j}$ is the cross-attention module. $|\boldsymbol{\Phi}_{d_j}|=8d^2+2d\times d_{\textrm{decffn}}$.

\begin{table*}[t]
\centering
\scalebox{0.75}{
\begin{tabular}{l|l|c|c|c|c}
\toprule
\toprule
\multirow{2}{*}{\textbf{Layer}}            & \multirow{2}{*}{\textbf{Module}} & \multirow{2}{*}{\textbf{\#params}} & \textbf{$d=512$}              & \textbf{$d=384$}             & \textbf{$d=768$}             \\
                                           &                                  &                                    & \multicolumn{1}{l|}{\#params / FLOPS} & \multicolumn{1}{l|}{\#params / FLOPS} & \multicolumn{1}{l}{\#params / FLOPS} \\ \midrule
\multirow{2}{*}{encoder layer}             & self-attn                        & $4d^2$                                & \multirow{2}{*}{3.15M / 95.4M}             & \multirow{2}{*}{1.77M / 53.9M}             & \multirow{2}{*}{7.08M / 214M}             \\
                                           & FFN                              & $8d^2$                                &                                    &                                    &                                    \\ \midrule
\multirow{3}{*}{vanilla decoder layer}     & self-attn                        & $4d^2$                                & \multirow{3}{*}{4.20M / 128M}             & \multirow{3}{*}{2.37M / 72.3M}             & \multirow{3}{*}{9.45M / 286M}             \\
                                           & cross-attn                       & $4d^2$                                &                                    &                                    &                                    \\
                                           & FFN                              & $8d^2$                                &                                    &                                    &                                    \\ \midrule
\multirow{3}{*}{interleaved decoder layer} & self-attn                        & $4d^2$                                & \multirow{3}{*}{2.23M / 72.9M}             & \multirow{3}{*}{1.25M / 41.3M}             & \multirow{3}{*}{5.01M / 162M}             \\
                                           & cross-attn                       & $4d^2$                                &                                    &                                    &                                    \\
                                           & 2 shared lightweight FFNs        & $d^2/2$                               &                                    &                                    &                                    \\ \bottomrule \bottomrule
\multicolumn{3}{l|}{\multirow{2}{*}{\textbf{Model}}}                                                                & \textbf{$d=512$}              & \textbf{$d=384$}             & \textbf{$d=768$}             \\
\multicolumn{3}{l|}{}                                                                                               & \#params / FLOPS                     & \#params / FLOPS                     & \#params / FLOPS                     \\ \midrule
\multicolumn{3}{l|}{6+6 Transformer (full parameterization)}                                                       &          44M / 1.84G                          &              25M / 1.13G                &          99M / 3.76G                          \\
\multicolumn{3}{l|}{12+2 Transformer (full parameterization)}                                                       &          46M / 1.90G                    &                   26M / 1.17G                &   104M / 3.89G                                \\
\multicolumn{3}{l|}{12+2 \textsc{Universal Transformer} (shared parameterization)}                                               &        7.4M / 1.90G                           &               4.1M / 1.17G                     &          16.5M / 3.89G                          \\
\multicolumn{3}{l|}{EdgeFormer (Ours)}                                                                         &          8.6M / 1.79G                         &             4.8M / 1.11G                 &      19.2M / 3.65G        \\ \bottomrule \bottomrule              
\end{tabular}
}
\caption{\textbf{Top:} \#params and FLOPS for Transformer layers. For the encoder and vanilla decoder layer, $d_{\textrm{ffn}}=4d$; while for the interleaved decoder layer, $d_{\textrm{ffn}}=d/4$. \textbf{Bottom:} \#params and FLOPS of whole models, where \#params excludes embedding lookup, and FLOPS is measured on a sample with src/tgt length of 30 and 32K vocabulary.}\label{tab:models}
\end{table*}

\subsection{Parameterization: Full vs Shared}
\paragraph{Full Parameterization}
Full parameterization is a common parameterization approach for Transformer, meaning that each model parameter (excluding embedding) is independent without being shared by multiple modules in the network. In a forward pass, each parameter is used only once. Full parameterization allows parameters to be flexible to fit their roles well during model training.

\paragraph{Shared Parameterization}
Despite the advantages of full parameterization, it is criticized to use large numbers of parameters inefficiently, motivating shared parameterization where multiple modules in a network share parameters.
For a model with shared parameterization (e.g., \textsc{Albert}), each model parameter is exploited more than once in a forward pass. The efficient parameterization can lead to a better result given the same parameterization budget, despite slowing down inference. On the other hand, given a fixed architecture (i.e., the same network depth and width), shared parameterization usually underperforms full parameterization because it has much fewer free model parameters.



\section{Constraints for On-device Seq2seq}\label{subsec:constraint}
\paragraph{Computation}
On-device computer vision (CV) models tend to use 1G FLOPS (0.5G MACS) as a constraint, which is directly followed by previous work on on-device translation~\citep{wu2020lite}. In our work, however, we propose to relax the FLOPS constraint for typical seq2seq tasks to \textbf{2G FLOPS (1G MACS)} because the latency requirement for on-device seq2seq generation is not so rigid as CV tasks and it is uncommon for an on-device seq2seq model to handle too many concurrent requests in practice. The relaxed constraint allows better prediction quality that strongly correlates with user experience for seq2seq tasks, but still ensure the CPU on edge devices to process tens of sentences per second, which is more than sufficient for an on-device seq2seq model. In addition to FLOPS that is a theoretical hardware-independent measurement for computational cost, we also require the runtime latency for an input sentence (typically $20\sim30$ tokens on average) to be within \textbf{around 100ms} using two mid-to-high-end CPU cores.


\paragraph{Memory}
In contrast to deploying a model on a cloud server without caring about memory cost much, there is a very strict memory constraint for an on-device model in practice, because a user's edge device (e.g., PC) is not only for model hosting; instead, it usually runs many other (background) apps and programs at the same time besides the model. To ensure moderate memory cost, we limit the number of model parameters (excluding word embedding lookup table) up to \textbf{10 million}, following previous work~\citep{wu2020lite}, and require the runtime memory footprint to be \textbf{less than 50MB}.





\section{EdgeFormer}
\subsection{Architecture}

The biggest challenge for an on-device seq2seq model is regarding the model size and memory cost. As shown in Table \ref{tab:models}, the number of parameters of a standard Transformer-base model ($d$=512) is about 45 million (excluding the embedding parameters), which is far beyond the parameterization budget (10 million) and unavoidably leads to massive memory cost despite acceptable FLOPS.

\begin{figure*}[t]
    \centering
    \begin{subfigure}[t]{0.5\textwidth}
        \centering
        \includegraphics[width=6.8cm]{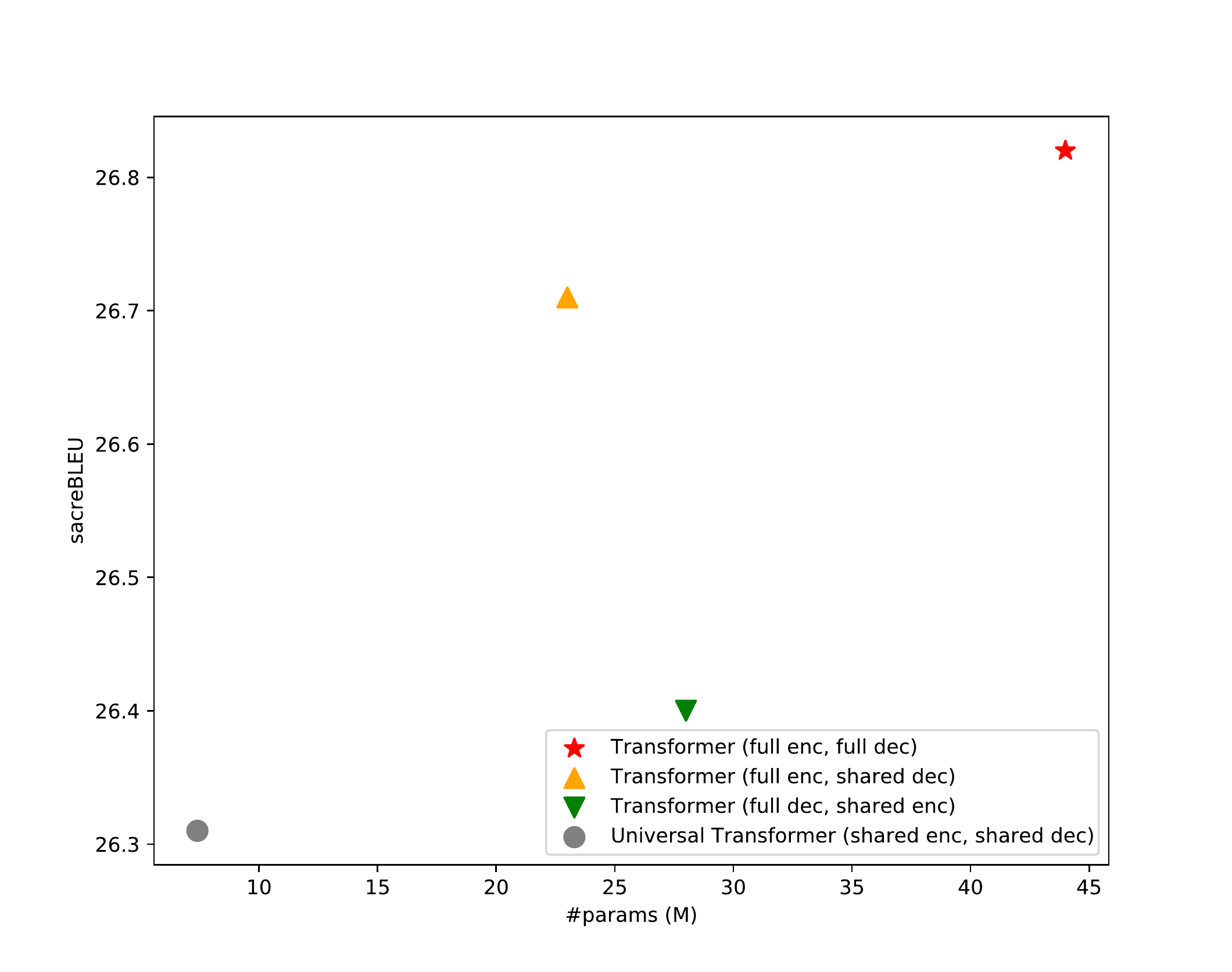}
        \caption{}\label{fig:encdec_param}
    \end{subfigure}%
    ~
    \begin{subfigure}[t]{0.5\textwidth}
        \centering
        \includegraphics[width=7cm]{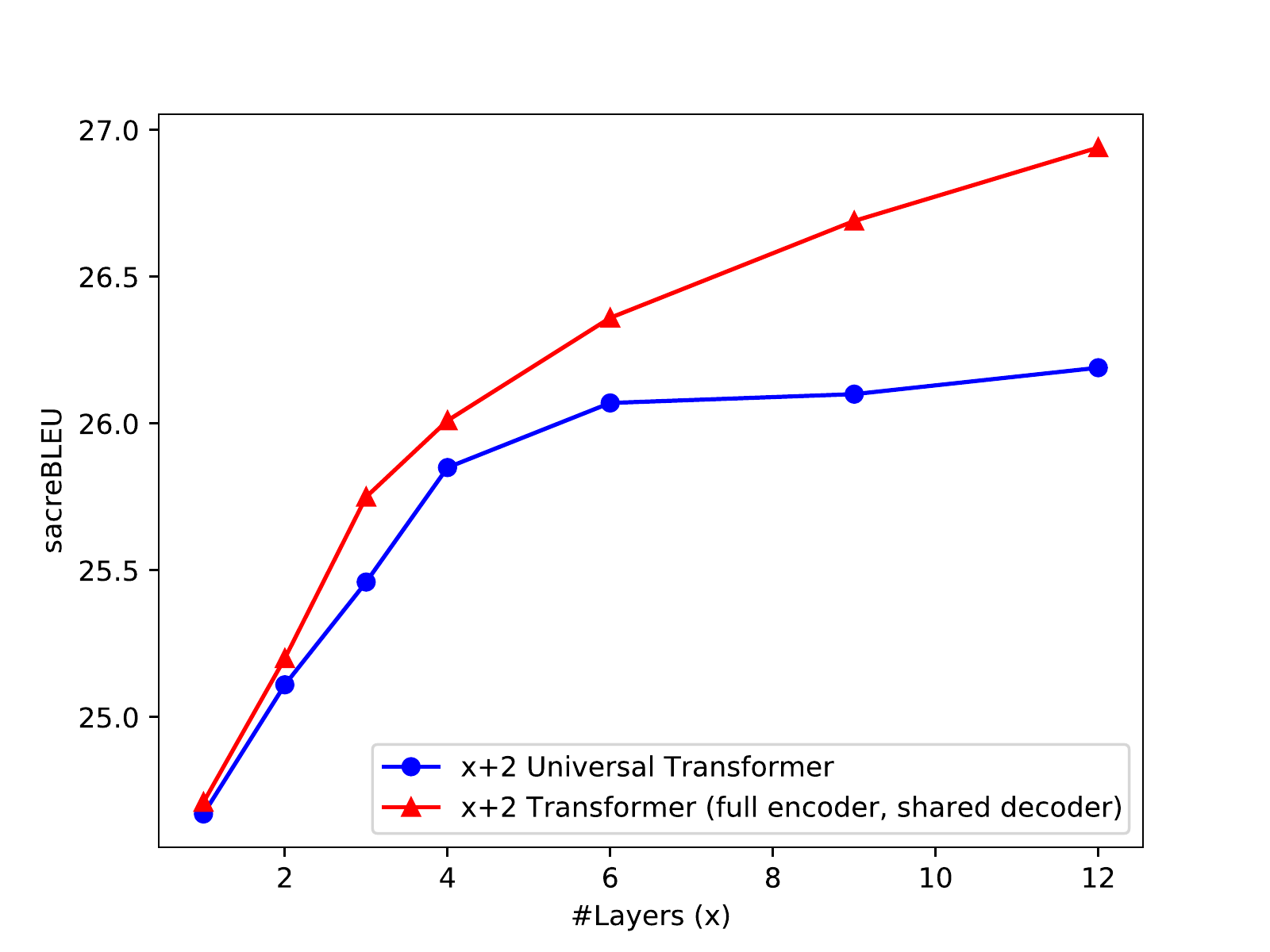}
        \caption{}\label{fig:LB}
    \end{subfigure}
    \caption{\textbf{(a)} Performance of 6+6 Transformer ($d=512$) on the newstest2013 English-German (En-De) translation dataset (dev set): densely parameterizing the decoder is  uneconomical and much less beneficial than parameterizing the encoder; \textbf{(b)} Comparison of x+2 Transformer with full-/shared-parameterized $x$ encoder layers on newstest2013 En-De dataset: when $x>6$, the performance of the Transformer with shared parameterization only improves marginally even if $x$ continues to increase.}
\end{figure*}

\textsc{EdgeFormer} is proposed to address the challenge. Instead of disruptive architectural changes\footnote{We basically follow the standard Transformer without major architectural changes because a standard Transformer should be more widely compatible and supported than a customized model architecture in user devices with various environments (e.g., hardware and runtime libraries).} as previous research~\citep{wu2020lite,mehta2020delight,panahi2021shapeshifter}, \textsc{EdgeFormer}'s architecture basically follows the standard Transformer consisting of a 12-layer encoder and 2-layer\footnote{We do not use 1-layer decoder because it does not consistently perform well~\citep{sun2021instantaneous}.} decoder, which is efficient in decoding. We mainly discuss the model with $d$=512 in this paper since it can achieve good performance in the on-device setting as long as it can be appropriately parameterized. The minor architectural modification we propose for \textsc{EdgeFormer} is using an interleaved decoder where attention modules are interleaved with shared lightweight\footnote{As observed by \citet{kasai2020deep}, reducing $d_{\textrm{decffn}}$ does not hurt the result much, as shown in Table \ref{tab:decffn} in Appendix \ref{sec:app_train}.} FFNs ($d_{\textrm{decffn}}<d$; in this work, $d_{\textrm{decffn}}=d/4$) in each decoder layer (shown in Figure \ref{fig:interleaved}). The modification is helpful for cost-effective parameterization (Section \ref{subsec:parameterization}):
\begin{itemize}
    \item The interleaved structure makes the architecture of encoder and decoder layers consistent~\citep{ma2021deltalm}, facilitating shared parameterization of attention modules throughout the encoder and decoder.
    \item As shown in Table \ref{tab:models}, the lightweight FFNs that interleave attention modules in the decoder reduce FLOPS and save a large number of parameters for decoder FFNs' parameterization that is very uneconomical. 
\end{itemize}




\subsection{Cost-effective Parameterization}\label{subsec:parameterization} 
Due to the tight parameterization budget (i.e., 10 million), \textsc{EdgeFormer} cannot be fully parameterized as in the standard way; instead, it has to adopt shared parameterization.

As a strong baseline for shared parameterization, \textsc{Universal Transformer} lets all its $M$ encoder layers share 1 group of encoder layer parameters and all its $N$ decoder layers share 1 group of decoder layer parameters:
\begin{displaymath}
\small
\boldsymbol{\Phi}_{e_1} \overset{\mathrm{tied}}{=} \boldsymbol{\Phi}_{e_2} \overset{\mathrm{tied}}{=} \cdots \overset{\mathrm{tied}}{=} \boldsymbol{\Phi}_{e_{M}}
\end{displaymath}
\begin{displaymath}
\small
\boldsymbol{\Phi}_{d_1} \overset{\mathrm{tied}}{=} \boldsymbol{\Phi}_{d_2} \overset{\mathrm{tied}}{=} \cdots \overset{\mathrm{tied}}{=} \boldsymbol{\Phi}_{d_{N}}
\end{displaymath}

Although \textsc{Universal Transformer} is a popular solution to shared parameterization, it is still not cost-effective for two reasons:


First, \textsc{Universal Transformer} uses (over) half of total parameters to parameterize the decoder, which is uneconomical. As shown in Figure \ref{fig:encdec_param}, given a fixed architecture (6+6 Transformer, $d=512$), densely parameterizing the decoder results in much less performance gain than parameterizing the encoder. This suggests we use as many parameters as possible to parameterize the encoder for the performance. 

Second, \textsc{Universal Transformer} does not consider load balance of model parameters, which was a rarely discussed problem until the recent emergence of Mixture-of-Expert models~\cite{fedus2021switch}. For the Transformers with a deep encoder and shallow decoder, \textsc{Universal Transformer}'s parameterization method will overburden parameters in the encoder but underutilize parameters in the decoder. For example, for a 12+2 \textsc{Universal Transformer}, a parameter in the encoder is used 12 times, while a parameter in the decoder is used only twice in a forward pass. As shown in Figure \ref{fig:LB}, moderately reusing parameters (e.g., when $x \le 4$) helps better utilize the parameters, resulting in significant performance gain without increasing parameters. However, as the shared parameters are overused (when $x>6$), the performance gain will become marginal, which is intuitive because a parameter's capability is limited. This suggests we balance the load of parameters to avoid them being either overused or underused.

Based on the above insights, we parameterize \textsc{EdgeFormer} in the following two novel principles for cost-effective parameterization:

\paragraph{Encoder-favored Parameterization}
For \textsc{EdgeFormer}, we parameterize its encoder using as many parameters as possible: except a small number of parameters ($d^2/2$) for all lightweight FFNs in the decoder, we use almost all parameters in our budget to parameterize the encoder. For attention modules in the decoder, we let them reuse (i.e., share) parameters with the attention modules in the encoder since attention modules in both the encoder and decoder work in the same mechanism and can be effectively shared~\citep{dong2019unified}. Thanks to the interleaved decoder architecture that makes the structure of encoder and decoder layers consistent, we let the self-attention module in a decoder layer share parameters with its corresponding odd layer in the encoder, and let its cross-attention module share with the corresponding even layer in the encoder, inspired by \citet{ma2021deltalm}:
\begin{displaymath}
\small
W^{[Q,K,V,O]}_{d_{j}} \overset{\mathrm{tied}}{=} W^{[Q,K,V,O]}_{e_{2j-1}} ~~~ (1 \le j \le 2)
\end{displaymath}
\begin{displaymath}
\small
W^{[\mathcal{Q},\mathcal{K},\mathcal{V},\mathcal{O}]}_{d_{j}} \overset{\mathrm{tied}}{=} W^{[Q,K,V,O]}_{e_{2j}} ~~~(1 \le j \le 2)
\end{displaymath}

\paragraph{Load-balanced Parameterization}
We try parameterizing \textsc{EdgeFormer} with a balanced load for each model parameter so that each parameter could be as equally exploited as possible in a forward pass. Given the parameterization budget and the load balance principle, we create 2 groups of encoder FFN parameters equally shared by all encoder layers, 1 group of decoder FFN parameters is shared by light FFNs in the decoder, and 4 groups of attention parameters are shared throughout the encoder and decoder. Except for parameters in the encoder FFNs that are used 6 times, other parameters are all used 4 times in a forward pass, resulting in a load balanced parameterization: 
\begin{displaymath}
\small
W^{[Q,K,V,O]}_{e_{i}} \overset{\mathrm{tied}}{=} W^{[Q,K,V,O]}_{e_{i+4}} ~~~ (1 \le i < 9)
\end{displaymath}
\begin{displaymath}
\small
W^{[f_1,f_2]}_{e_{i}} \overset{\mathrm{tied}}{=} W^{[f_1,f_2]}_{e_{i+2}} ~~~ (1 \le i < 11)
\end{displaymath}
\begin{displaymath}
\small
W^{[f_1,f_2]}_{d_{j}} \overset{\mathrm{tied}}{=} W^{[f_1,f_2]}_{d_1} ~~~ (1 \le j \le 2)
\end{displaymath}

\subsection{Layer Adaptation}
Shared parameterization causes layers with tied weights to become less specialized, as discussed in Section \ref{sec:intro}. To allow tied layers to be better adapted to their corresponding roles, we propose layer adaptation to further enhance \textsc{EdgeFormer}. Inspired by parameter-efficient task transfer methods, we investigate three efficient layer adaption approaches:   

\paragraph{Bias-based Layer Adaptation (Bias-LA)}
Inspired by BitFit~\citep{ben2021bitfit} fine-tuning with only bias terms, we untie all bias terms of each layer and use them to specialize the layers with tied weights, as shown in Figure \ref{fig:full_LA}(b). As BitFit, bias-based layer adaptation introduces very few additional parameters without inference overhead.

\paragraph{Adapter-based Layer Adaptation (Adapter-LA)}
Adapter-based approaches~\citep{houlsby2019parameter} introduce adapter modules for NLP task transfer without full fune-tuning. We borrow this idea for layer adaptation by introducing an independent adapter module for each layer. Specifically, we adopt the recently proposed LoRA~\citep{hu2021lora} as our layer adapter, as Figure \ref{fig:full_LA}(c) shows. In our experiments, we apply the layer adapter to $W^Q$ and $W^V$, as the original paper of LoRA suggests.

\paragraph{Prefix-based Layer Adaptation (Prefix-LA)}
Inspired by recent work~\citep{li2021prefix,lester2021power} using a prefix/prompt for task transfer, we introduce $L$ tokens with learnable parameters as a specific prefix for each layer to adapt layers with tied weights, as shown in Figure \ref{fig:full_LA}(d). The prefixs are only used for keys and values in attention modules, which will not introduce much inference overhead as long as $L$ is moderately set.

Following the encoder-favored principle in Section \ref{subsec:parameterization}, we only apply LA to encoder layers.

\begin{table*}[t]
\centering
\small
\begin{tabular}{l|c|c|c}
\toprule
\multicolumn{1}{c|}{\textbf{Model}} & \textbf{\#Params} & \textbf{FLOPS} & \textbf{sacreBLEU} \\ \midrule
Teacher                            & 176M              & 6.7G             & 29.3             \\ \midrule
6+6 Transformer (full enc, full dec)         & 44M               & 1.8G             & 28.5               \\ 
6+6 Transformer (full enc, shared dec)          & 23M               & 1.8G             & 28.2               \\ 
6+6 Transformer (full dec, shared enc)      & 28M               & 1.8G             & 27.3               \\  \midrule
12+2 Transformer (full enc, full dec)         & 46M               & 1.9G             & 28.5               \\ 
12+2 Transformer (full enc, shared dec)          & 42M               & 1.9G             & 28.4               \\ 
12+2 Transformer (full dec, shared enc)      & 12M               & 1.9G             & 27.2               \\  \midrule
12+2 UT        & 7.4M              & 1.9G             & 27.0               \\ 
12+2 UT ($d_\textrm{ffn}=2560$)         & 8.5M              & 2.1G             & 27.2               \\
12+2 UT ($d_\textrm{encffn}=3072$)$^1$         & 8.5M              & 2.3G             & 27.4               \\
12+2 UT ($d_\textrm{decffn}=3072$)         & 8.5M              & 2.0G             & 27.0               \\
\midrule       
\textsc{EdgeFormer} w/o LA$^2$                & 8.6M              & 1.8G             & 27.7$^{\dag(1)}$\\    
\textsc{EdgeFormer} (Bias-LA)        & 8.6M              & 1.8G             & 27.8  \\  
\textsc{EdgeFormer} (Adapter-LA) ($r=32$)  & 9.4M              & 1.8G             & 28.0$^{\dag(2)}$ \\  
\textsc{EdgeFormer} (Prefix-LA) ($L=8$)   & 8.6M              & 1.9G             & 28.0$^{\dag(2)}$  \\  \bottomrule
\end{tabular}
\caption{WMT14 En-De results. To fairly compare with \textsc{Universal Transformer} (UT) that is originally smaller than \textsc{EdgeFormer}, we also test UT with larger FFNs to make its model size comparable to \textsc{EdgeFormer}. $^{\dag(i)}$ denotes $p<0.05$ in significance test compared with the model marked with $^i$.}\label{tab:wmt14ende}
\end{table*}

\begin{table*}[t!]
\centering
\small
\begin{tabular}{l|c|c|c|c}
\toprule
\multicolumn{1}{c|}{\textbf{FFNs}} & \textbf{Load} & \textbf{\#Params} & \textbf{FLOPS} & \textbf{sacreBLEU} \\ \midrule
2 FFNs ($d_{\textrm{ffn}}=2048$)$^1$                            & 6-6           & 8.6M              & 1.8G           & 27.7               \\ \midrule
3 FFNs ($d_{\textrm{ffn}}=1536$)                           & 4-4-4         & 9.1M              &   1.6G             &        27.4$^{\dag(1)}$            \\
4 FFNs ($d_{\textrm{ffn}}=1024$)                           & 3-3-3-3       & 8.6M              &     1.4G          &         27.2$^{\dag(1)}$            \\ \midrule
2 FFNs ($d_{\textrm{ffn}}=2048$)                          & 1-11          & 8.6M              & 1.8G       &    27.5$^{\dag(1)}$                \\
2 FFNs ($d_{\textrm{ffn}}=2048$)                           & 11-1          & 8.6M              & 1.8G           &    27.4$^{\dag(1)}$               \\ \bottomrule
\end{tabular}
\caption{Performance of \textsc{EdgeFormer} with various encoder FFN parameterization on WMT14 En-De. Load 6-6 means the 2 groups of FFN parameters are used 6 times each, while Load 1-11 means 1 group of FFN is used once, and the other is used 11 times.}\label{tab:load_exp}
\end{table*}

\section{Experiments}
\subsection{Experimental Setting}
We mainly evaluate our approach in Machine Translation (MT). We select the most popular MT benchmark -- WMT14 English-German (En-De) translation task, which is also a touchstone for seq2seq evaluation, as our main test bed. To compare with previous work, we also evaluate WMT14 English-French (En-Fr) translation. We follow the standard way to train and evaluate evaluate WMT14 En-De and En-Fr. As \citet{ott2018scaling}, we use a joint source-target dictionary of 32K Byte Pair Encoding (BPE) for En-De, and 40K BPE for En-Fr. We mainly use sacreBLEU~\citep{post2018call} for evaluation.


We select \textsc{Universal Transformer} which is the most popular and a strong baseline of parameter-efficient Transformer for fair comparison. By default, we apply Seq-KD~\citep{kim2016sequence} to train models and use the full-parameterized 6+6 Transformer-big ($d=1,024$) model~\citep{vaswani2017attention, ott2018scaling} as the teacher.

By default, for each experiment, we train 5 models with different initializations and report their average evaluation results for Table \ref{tab:wmt14ende}, \ref{tab:load_exp} and \ref{tab:pretrain} with significance test. For inference, we use beam search (beam=5).

\subsection{Offline Evaluation}
We evaluate \textsc{EdgeFormer} and compare it with \textsc{Universal Transformer} (UT) on WMT14 En-De. According to Table \ref{tab:wmt14ende}, the \textsc{EdgeFormer} without layer adaptation (LA) largely outperforms UTs. Among the LA approaches, both Adapter-LA and Prefix-LA are clear to benefit the result with marginal computational or parameterization cost, while Bias-LA does not show significant performance gain though it is the cheapest.

\begin{figure*}[t]
    \centering
    \includegraphics[width=14cm]{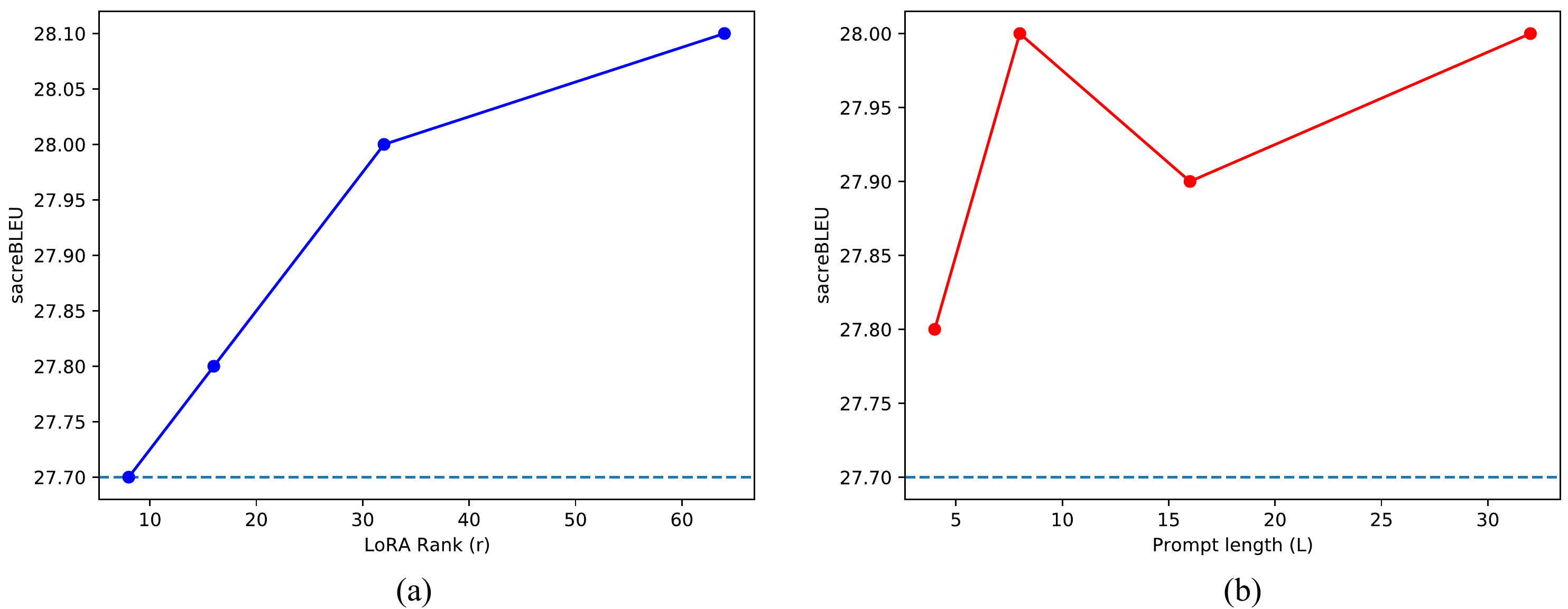}
    \caption{The effects of \textbf{(a)} rank $r$ in Adapter-LA, and \textbf{(b)} prefix length $L$ in Prefix-LA on the performance in WMT14 En-De. Note that $r=64$ will lead to exceed our parameterization budget despite better performance.}
    \label{fig:la_res}
\end{figure*}

\begin{table*}[t!]
\centering
\small
\begin{tabular}{l|c|c|c|c}
\toprule
\multicolumn{1}{c|}{\textbf{Model}} & \textbf{\#params} & \textbf{FLOPS} & \textbf{En-De} & \textbf{En-Fr}\\ \midrule
6+6 Transformer                    & \underline{44M}               & 1.8G           & 28.3                 & 41.0               \\
12+2 Transformer                   & \underline{46M}               & 1.9G           & 28.4                 & 41.4                  \\
12+2 UT                            & 7.4M              & 1.9G           & 26.2 & 39.2                 \\ \midrule
DeLighT                            & \underline{31.4M}             & -              & \bf 27.6                 & 39.6                 \\
Shapeshifter                       & 8.2M              & -              & 26.6                 & \bf 40.8                \\
Lite Transformer (small)                   & 2.9M             & 0.2G           & 22.5                 & 35.3 \\
Lite Transformer (medium)                   & \underline{11.7M}             & 0.7G           & 25.6                 & 39.1 \\
Lite Transformer (big)                   & \underline{17.3M}             & 1.0G           & 26.5                 & 39.6                 \\ \midrule
EdgeFormer w/o LA                  & 8.6M              & 1.8G           & 26.5                 & 39.8                \\
EdgeFormer (Adapter-LA)       & 9.4M              & 1.8G           & \bf 26.9                 & \bf 40.5                \\
EdgeFormer (Prefix-LA)         & 8.6M              & 1.9G           & 26.8                 & 40.3                \\ \bottomrule
\end{tabular}
\caption{Result comparison to previous parameter-efficient Transformers that have fewer parameters than the baseline Transformer (around 45M parameters). ``-'' means that the metrics are unavailable or not comparable in the original paper. The underlines denote that the metrics cannot meet the on-device requirement. Note that all the models in this table do not apply Seq-KD.\label{tab:sota}}
\end{table*}

As discussed in Section \ref{subsec:parameterization}, the advantage of \textsc{EdgeFormer} over UT comes from its cost-effective parameterization. The encoder-favored principle is again supported by comparing 6+6 Transformers' results in Table \ref{tab:wmt14ende}, which is consistent with the observation on the dev set in Figure \ref{fig:encdec_param}. To further understand the effectiveness of load-balanced parameterization principles, we conduct an ablation study by adjusting encoder FFNs in \textsc{EdgeFormer}. Table \ref{tab:load_exp} shows the results of \textsc{EdgeFormer} with various FFN parameterization. As we reduce $d_{\textrm{ffn}}$ (e.g., to 1536 or 1024), we can increase the group of encoder FFN parameters and reduce their load given a fixed parameterization budget. However, such a strategy leads to a clear degradation of sacreBLEU. One reason is that the FFN parameters of a reduced load (3-4 times) are not so fully utilized as the baseline (6 times) despite other reasons such as the differences of network shape (e.g., $d_{\textrm{ffn}}$). To minimize the effects of other factors, we compare the first group with a balanced parameter load (i.e., 6-6) and the last group with a imbalanced parameter load (1-11 or 11-1), showing load-balanced parameterization is consistently better than the imbalanced counterparts.

After discussing parameterization, we then analyze the effects of layer adaptation on the results by mainly focusing on Adapter-LA and Prefix-LA that both show performance gain. Figure \ref{fig:la_res} shows the effects of the rank $r$ in Adapter-LA and prefix length $L$ in Prefix-LA. As $r$ increases, the model performance will gradually improve. However, when $r$ becomes large (e.g., $r \ge 64$), it will exceed our parameterization budget and thus the gain will become meaningless. As for prefix length $L$ in Prefix-LA, it is different from $r$ that it will not keep improving the results as it increases: the gain can hardly be observed after some length (e.g., $L=8$), which is similar to the observation in prefix-tuning~\citep{li2021prefix}. Therefore, we use $r=32$ and $L=8$ as the default setting to report the results of Adapter-LA and Prefix-LA.

\begin{table*}[t]
\centering
\small
\scalebox{1}{
\begin{tabular}{c|c|c|c|c|c} \toprule
\multicolumn{6}{c}{\textbf{WMT14 En-De}}                                                                                                            \\ \midrule
\textbf{Model}                    & \textbf{Disk size (7zip)} & \textbf{Peak Memory} & \textbf{Latency 1} & \textbf{Latency 2} & \textbf{sacreBLEU} \\ \midrule
EdgeFormer (Adapter-LA, 32k vocab) & 28MB                      & 60MB                 & 65ms               & 114ms              & 27.2               \\
EdgeFormer (Adapter-LA, 8k vocab)  & 15MB                      & 47MB                 & 59ms               & 101ms              & 27.1               \\ \midrule
\multicolumn{6}{c}{\textbf{CoNLL-14}}                                                                                                               \\ \midrule
\textbf{Model}                    & \textbf{Disk size (7zip)} & \textbf{Peak Memory} & \textbf{Latency 1} & \textbf{Latency 2} & $F_{0.5}$     \\ \midrule
EdgeFormer (Adapter-LA, 2k vocab)  & 11MB                      & 42MB                 & 51ms               & 98ms               & 50.8              \\ \bottomrule
\end{tabular}}
\caption{Runtime results for int8-quantized \textsc{EdgeFormer}, in which Latency1 and Latency 2 denote the average latency per sentence measured on the Intel® Xeon® E-2288G CPU and Qualcomm SM8150 Snapdragon 855 CPU, respectively. We run through the test set with batch size=1, and use greedy decoding instead of beam search.}\label{tab:runtime}
\end{table*}

\begin{table*}[t]
\centering
\small
\scalebox{0.9}{
\begin{tabular}{l|c|c|c|ccc|ccc}
\toprule
\multicolumn{1}{c|}{\multirow{2}{*}{\textbf{Model}}} & \multirow{2}{*}{\textbf{\#Param}} & \multirow{2}{*}{\textbf{FLOPS}} & \textbf{CoNLL14} & \multicolumn{3}{c|}{\textbf{XSum}} & \multicolumn{3}{c}{\textbf{SQuAD-NQG}} \\
\multicolumn{1}{c|}{}                                &                                   &                                 & $F_{0.5}$             & RG-1      & RG-2      & RG-L      & B4        & MTR      & RG-L     \\  \midrule
Transformer-base                                    & 44M                               & 1.8G                            & 50.1            & 31.2      & 10.7      & 24.9      &      2.6$^*$     &   9.0$^*$   &     26.0$^*$     \\ \midrule
Pretrained 12+2 UT ($d_{ffn}=2048$)                                 & 7.4M                              & 1.4G                            & 50.8             & 36.0 &      14.5     &      29.2 &     19.8      &      22.2    &   46.9       \\
Pretrained 12+2 UT ($d_{ffn}=3072$)$^{1}$                           & 9.4M                              & 1.9G                            & 51.1             & 36.7      & 14.9      & 29.7      &      20.1     &      22.4    &  47.1        \\ \midrule
\textsc{EdgeLM}              & 9.4M                              & 1.3G                            & \bf 52.0$^{(1)}$             & \bf 37.2$^{(1)}$      & \bf 15.4$^{(1)}$      & \bf 30.3$^{(1)}$      & \bf 20.6$^{(1)}$      & \bf 23.0$^{(1)}$     & \bf 47.4$^{(1)}$     \\ \bottomrule
\end{tabular}
}
\caption{The performance of \textsc{EdgeLM} in comparison with the baselines. $*$ denotes that the results are from \citet{chen2019reinforcement}.}\label{tab:pretrain}
\end{table*}

Finally, we compare \textsc{EdgeFormer} with recent work on parameter-efficient Transformer modeling. To keep consistency of the training and evaluation protocols with previous work, we here give up using Seq-KD to train the models, and report BLEU~\citep{papineni2002bleu} for comparison. Specifically, we compare with DeLighT~\citep{mehta2020delight}, Shapeshifter~\citep{panahi2021shapeshifter} and Lite Transformer~\citep{wu2020lite}, and show the results in Table \ref{tab:sota}. However, it is notable that the results are not strictly comparable because the previous studies have their own focus and setting, which are different from ours. For example, DeLighT and Lite Transformer focus much more on FLOPS than the model size, thus they do not a desirable tradeoff between the model quality and size; while Shapeshifter's goal is minimizing the model size despite an additional $10\%\sim20\%$ inference overhead. Regardless of these factors that prevent fair comparison, \textsc{EdgeFormer} achieves 26.9 BLEU in En-De under the strict on-device resource constraints, which outperforms the state-of-the-art Shapeshifter with the similar model size despite. It is notable that \textsc{EdgeFormer} here uses the same model architecture and training configuration for both En-De and En-Fr, while Shapeshifter uses different model architecture configurations specific for En-De and En-Fr, which may account for its better performance in En-Fr.



\subsection{Runtime Evaluation}
We conduct experiments in WMT14 En-De translation and CoNLL-14 Grammatical Error Correction\footnote{We include experiments details of GEC in Appendix \ref{sec:app_train}.} (GEC) benchmark for runtime latency and memory evaluation using onnxruntime\footnote{\url{https://github.com/microsoft/onnxruntime}} that supports efficient seq2seq decoding. We apply int8-quantization to \textsc{EdgeFormer} and test latency on 2 devices: a 2-core Intel® Xeon® E-2288G CPU (in PC), and a 2-core Qualcomm SM8150 Snapdragon 855 CPU (in Pixel 4), which are both current mid-to-high end CPUs launched 2-3 years ago.

Table \ref{tab:runtime} shows runtime evaluation results. 
With int8-quantization and smaller vocabulary, \textsc{EdgeFormer} can not only meet the on-device seq2seq requirements but also maintain its good performance, demonstrating its practical values.

\section{EdgeLM -- The Pretrained EdgeFormer}
Given the promising results, we introduce \textsc{EdgeLM} -- the pretrained\footnote{We include pretraining details in the Appendix \ref{sec:app_pretrain}.} \textsc{EdgeFormer} (Adapter-LA) with 8K sentenpiece vocabulary with factorized embedding ($d_{\textrm{embed}}=128$) through the same self-supervised task (i.e., masked span infilling) as T5~\citep{raffel2019exploring} and make it publicly available for downstream on-device seq2seq task fine-tuning.

We evaluate \textsc{EdgeLM} in the benchmarks of three popular seq2seq tasks: CoNLL-14 for Grammatical Error Correction (GEC), XSum~\citep{narayan2018don} for Abstractive Summarization, and SQuAD-NQG~\citep{du2017learning} for Question Generation. According to Table \ref{tab:pretrain}, \textsc{EdgeLM} achieves significantly better performance than the pretrained UT models as well as the Transformer-base model trained from scratch. We believe that \textsc{EdgeLM}, as the first publicly released on-device seq2seq pretrained model, can largely facilitate on-device seq2seq generation in practice.

\section{Related Work}
On-device seq2seq generation in NLP is a research area that has been less explored than on-device CV and NLU~\cite{tambe2021edgebert}. Besides the general techniques like pruning, compression, quantization and knowledge distillation~\cite{fan2019reducing,xu2020bert,li2022dq} that are orthogonal to our effort, parameter-efficient Transformer-based seq2seq modeling is the most related research branch to ours. In this branch, \textsc{Universal Transformer}~\citep{dehghani2018universal} uses cross-layer sharing method, which is the most popular solution to parameter efficiency. \citet{takase2021lessons} extends \textsc{Universal Transformer} by studying different ways for layer sharing, and \citet{reid2021subformer} proposes to free the first and last encoder layer and widen the intermediate layers for better performance. However, both the approaches consider parameter-efficiency only without caring about latency becoming worse. 

In addition to work improving parameter efficiency by weight sharing, there is research that studies lightweight model architecture for seq2seq learning where early work~\citep{gehring2017convolutional,wu2018pay} mainly focuses on CNNs, while recent efforts have tended to switch to attention-based models such as \citet{mehta2020delight}. Also, low-rank factorization has been studied intensively to make the model tiny~\citep{zhang2021beyond,panahi2021shapeshifter}; and hardware-aware network architecture search with elastic modeling~\citep{wang2020hat} has been proposed recently for facilitating deployment of seq2seq models on various devices. Among previous studies, the work of \citet{wu2020lite} is the most related to ours, which studies seq2seq generation in an on-device setting. However, it sets the computational constraint for on-device seq2seq to be the same with the CV tasks, which is too strict and unnecessary, as discussed in Section \ref{subsec:constraint}. As a result, their models focus on FLOPS optimization much more than memory, leading to an undesirable tradeoff between the quality and model size for the practical on-device seq2seq setting which should care about memory much more than latency. In contrast, our work carefully evaluates bottleneck constraints, and proposes appropriate models with parameterization and layer adaptation innovations, largely improving the results for practical on-device seq2seq generation.

\section{Conclusion and Future Work}\label{sec:future}
We formally study on-device seq2seq generation, including defining its practical resource constraint setting and proposing an appropriate modeling technology \textsc{EdgeFormer}. The cost-effective parameterization and layer adaptation innovations in \textsc{EdgeFormer} both prove effective to improve the results with negligible computation and memory cost, achieving state-of-the-art results in the on-device seq2seq generation setting. Our released pretrained \textsc{EdgeFormer} -- \textsc{EdgeLM} can be easily fine-tuned for downstream seq2seq tasks, largely facilitating on-device seq2seq generation in practice.

For future work, we plan to further study load-balanced parameterization for parameter-efficient models, which is an interesting and new but seemingly profound machine learning research problem: instead of naively assuming that all the parameters are equal in this preliminary study, we suspect that parameters in different modules (e.g., parameters in the self-attn and FFN; or parameters in different layers) should be under different amounts of load. We look forward to in-depth research on this problem, which might be helpful to deepen our understanding of neural networks.


\section{Limitations}
\textsc{EdgeFormer} is a preliminary model proposed for on-device seq2seq generation setting, which still has much room for improvement. For example, as mentioned in Section \ref{sec:future}, the current load balance mechanism naively assumes that the number of times that a parameter is used in a forward pass is equal to its load, which may not be always true because parameters in different moduels are different: some parameters may be effectively used more times than others, which requires deeper understanding of neural network and the Transformer.

\section*{Acknowledgments}
We thank all the anonymous reviewers for their valuable comments. We thank Xiaohu Tang, Fucheng Jia, Yifan Yang and Huiqiang Jiang for their help in runtime evaluation, and thank Shuming Ma, Ting Cao, Fan Yang, Qiufeng Yin, Yuqing Yang and Lidong Zhou in Microsoft Research Asia for the discussion and helpful comments. We also appreciate the support from Wenbing Li, Yufeng Li, Bowen Bao, Ye Wang, Sunghoon Choi, Scott McKay and Emma Ning in Microsoft AI Frameworks for onnxruntime, and appreciate the feedback and valuable suggestions from Joshua Burkholder, Xun Wang, Weixin Cai and Zhang Li in Microsoft Office Intelligence regarding detailed constraints for on-device seq2seq generation in real-word applications.

\bibliography{edgeformer}

\begin{thebibliography}{44}
\providecommand{\natexlab}[1]{#1}
\providecommand{\url}[1]{\texttt{#1}}
\expandafter\ifx\csname urlstyle\endcsname\relax
  \providecommand{\doi}[1]{doi: #1}\else
  \providecommand{\doi}{doi: \begingroup \urlstyle{rm}\Url}\fi

\bibitem[Ben~Zaken et~al.(2021)Ben~Zaken, Ravfogel, and
  Goldberg]{ben2021bitfit}
Elad Ben~Zaken, Shauli Ravfogel, and Yoav Goldberg.
\newblock Bitfit: Simple parameter-efficient fine-tuning for transformer-based
  masked language-models.
\newblock \emph{arXiv e-prints}, pp.\  arXiv--2106, 2021.

\bibitem[Bryant et~al.(2019)Bryant, Felice, Andersen, and
  Briscoe]{bryant2019bea}
Christopher Bryant, Mariano Felice, {\O}istein~E Andersen, and Ted Briscoe.
\newblock The bea-2019 shared task on grammatical error correction.
\newblock In \emph{Proceedings of the Fourteenth Workshop on Innovative Use of
  NLP for Building Educational Applications}, pp.\  52--75, 2019.

\bibitem[Chen et~al.(2020)Chen, Ge, Zhang, Wei, and
  Zhou]{chen-etal-2020-improving-efficiency}
Mengyun Chen, Tao Ge, Xingxing Zhang, Furu Wei, and Ming Zhou.
\newblock Improving the efficiency of grammatical error correction with
  erroneous span detection and correction.
\newblock In \emph{Proceedings of the 2020 Conference on Empirical Methods in
  Natural Language Processing (EMNLP)}, pp.\  7162--7169. Association for
  Computational Linguistics, 2020.

\bibitem[Chen et~al.(2019)Chen, Wu, and Zaki]{chen2019reinforcement}
Yu~Chen, Lingfei Wu, and Mohammed~J Zaki.
\newblock Reinforcement learning based graph-to-sequence model for natural
  question generation.
\newblock \emph{arXiv preprint arXiv:1908.04942}, 2019.

\bibitem[Dahlmeier \& Ng(2012)Dahlmeier and Ng]{dahlmeier2012better}
Daniel Dahlmeier and Hwee~Tou Ng.
\newblock Better evaluation for grammatical error correction.
\newblock In \emph{Proceedings of the 2012 Conference of the North American
  Chapter of the Association for Computational Linguistics: Human Language
  Technologies}, pp.\  568--572, 2012.

\bibitem[Dehghani et~al.(2018)Dehghani, Gouws, Vinyals, Uszkoreit, and
  Kaiser]{dehghani2018universal}
Mostafa Dehghani, Stephan Gouws, Oriol Vinyals, Jakob Uszkoreit, and {\L}ukasz
  Kaiser.
\newblock Universal transformers.
\newblock \emph{arXiv preprint arXiv:1807.03819}, 2018.

\bibitem[Dhar et~al.(2019)Dhar, Guo, Liu, Tripathi, Kurup, and
  Shah]{dhar2019device}
Sauptik Dhar, Junyao Guo, Jiayi Liu, Samarth Tripathi, Unmesh Kurup, and Mohak
  Shah.
\newblock On-device machine learning: An algorithms and learning theory
  perspective.
\newblock \emph{arXiv preprint arXiv:1911.00623}, 2019.

\bibitem[Dong et~al.(2019)Dong, Yang, Wang, Wei, Liu, Wang, Gao, Zhou, and
  Hon]{dong2019unified}
Li~Dong, Nan Yang, Wenhui Wang, Furu Wei, Xiaodong Liu, Yu~Wang, Jianfeng Gao,
  Ming Zhou, and Hsiao-Wuen Hon.
\newblock Unified language model pre-training for natural language
  understanding and generation.
\newblock \emph{arXiv preprint arXiv:1905.03197}, 2019.

\bibitem[Du et~al.(2017)Du, Shao, and Cardie]{du2017learning}
Xinya Du, Junru Shao, and Claire Cardie.
\newblock Learning to ask: Neural question generation for reading
  comprehension.
\newblock In \emph{Association for Computational Linguistics (ACL)}, 2017.

\bibitem[Fan et~al.(2019)Fan, Grave, and Joulin]{fan2019reducing}
Angela Fan, Edouard Grave, and Armand Joulin.
\newblock Reducing transformer depth on demand with structured dropout.
\newblock In \emph{International Conference on Learning Representations}, 2019.

\bibitem[Fedus et~al.(2021)Fedus, Zoph, and Shazeer]{fedus2021switch}
William Fedus, Barret Zoph, and Noam Shazeer.
\newblock Switch transformers: Scaling to trillion parameter models with simple
  and efficient sparsity, 2021.

\bibitem[Gehring et~al.(2017)Gehring, Auli, Grangier, Yarats, and
  Dauphin]{gehring2017convolutional}
Jonas Gehring, Michael Auli, David Grangier, Denis Yarats, and Yann~N Dauphin.
\newblock Convolutional sequence to sequence learning.
\newblock In \emph{International Conference on Machine Learning}, pp.\
  1243--1252. PMLR, 2017.

\bibitem[Houlsby et~al.(2019)Houlsby, Giurgiu, Jastrzebski, Morrone,
  De~Laroussilhe, Gesmundo, Attariyan, and Gelly]{houlsby2019parameter}
Neil Houlsby, Andrei Giurgiu, Stanislaw Jastrzebski, Bruna Morrone, Quentin
  De~Laroussilhe, Andrea Gesmundo, Mona Attariyan, and Sylvain Gelly.
\newblock Parameter-efficient transfer learning for nlp.
\newblock In \emph{International Conference on Machine Learning}, pp.\
  2790--2799. PMLR, 2019.

\bibitem[Hu et~al.(2021)Hu, Shen, Wallis, Allen-Zhu, Li, Wang, Wang, and
  Chen]{hu2021lora}
Edward~J Hu, Yelong Shen, Phillip Wallis, Zeyuan Allen-Zhu, Yuanzhi Li, Shean
  Wang, Lu~Wang, and Weizhu Chen.
\newblock Lora: Low-rank adaptation of large language models.
\newblock \emph{arXiv preprint arXiv:2106.09685}, 2021.

\bibitem[Kasai et~al.(2020)Kasai, Pappas, Peng, Cross, and
  Smith]{kasai2020deep}
Jungo Kasai, Nikolaos Pappas, Hao Peng, James Cross, and Noah~A Smith.
\newblock Deep encoder, shallow decoder: Reevaluating non-autoregressive
  machine translation.
\newblock \emph{arXiv preprint arXiv:2006.10369}, 2020.

\bibitem[Kim \& Rush(2016)Kim and Rush]{kim2016sequence}
Yoon Kim and Alexander~M Rush.
\newblock Sequence-level knowledge distillation.
\newblock \emph{arXiv preprint arXiv:1606.07947}, 2016.

\bibitem[Kudo \& Richardson(2018)Kudo and Richardson]{kudo2018sentencepiece}
Taku Kudo and John Richardson.
\newblock Sentencepiece: A simple and language independent subword tokenizer
  and detokenizer for neural text processing.
\newblock \emph{arXiv preprint arXiv:1808.06226}, 2018.

\bibitem[Lester et~al.(2021)Lester, Al-Rfou, and Constant]{lester2021power}
Brian Lester, Rami Al-Rfou, and Noah Constant.
\newblock The power of scale for parameter-efficient prompt tuning.
\newblock \emph{arXiv preprint arXiv:2104.08691}, 2021.

\bibitem[Li \& Liang(2021)Li and Liang]{li2021prefix}
Xiang~Lisa Li and Percy Liang.
\newblock Prefix-tuning: Optimizing continuous prompts for generation.
\newblock \emph{arXiv preprint arXiv:2101.00190}, 2021.

\bibitem[Li et~al.(2022)Li, Wang, Tan, Nallapati, Bhatia, Arnold, Xiang, and
  Roth]{li2022dq}
Zheng Li, Zijian Wang, Ming Tan, Ramesh Nallapati, Parminder Bhatia, Andrew
  Arnold, Bing Xiang, and Dan Roth.
\newblock Dq-bart: Efficient sequence-to-sequence model via joint distillation
  and quantization.
\newblock \emph{arXiv preprint arXiv:2203.11239}, 2022.

\bibitem[Liu et~al.(2019)Liu, Ott, Goyal, Du, Joshi, Chen, Levy, Lewis,
  Zettlemoyer, and Stoyanov]{liu2019roberta}
Yinhan Liu, Myle Ott, Naman Goyal, Jingfei Du, Mandar Joshi, Danqi Chen, Omer
  Levy, Mike Lewis, Luke Zettlemoyer, and Veselin Stoyanov.
\newblock Roberta: A robustly optimized bert pretraining approach.
\newblock \emph{arXiv preprint arXiv:1907.11692}, 2019.

\bibitem[Ma et~al.(2021)Ma, Dong, Huang, Zhang, Muzio, Singhal, Awadalla, Song,
  and Wei]{ma2021deltalm}
Shuming Ma, Li~Dong, Shaohan Huang, Dongdong Zhang, Alexandre Muzio, Saksham
  Singhal, Hany~Hassan Awadalla, Xia Song, and Furu Wei.
\newblock Deltalm: Encoder-decoder pre-training for language generation and
  translation by augmenting pretrained multilingual encoders.
\newblock \emph{arXiv preprint arXiv:2106.13736}, 2021.

\bibitem[Mehta et~al.(2020)Mehta, Ghazvininejad, Iyer, Zettlemoyer, and
  Hajishirzi]{mehta2020delight}
Sachin Mehta, Marjan Ghazvininejad, Srinivasan Iyer, Luke Zettlemoyer, and
  Hannaneh Hajishirzi.
\newblock Delight: Deep and light-weight transformer.
\newblock \emph{arXiv preprint arXiv:2008.00623}, 2020.

\bibitem[Mizumoto et~al.(2011)Mizumoto, Komachi, Nagata, and
  Matsumoto]{mizumoto2011mining}
Tomoya Mizumoto, Mamoru Komachi, Masaaki Nagata, and Yuji Matsumoto.
\newblock Mining revision log of language learning sns for automated japanese
  error correction of second language learners.
\newblock In \emph{Proceedings of 5th International Joint Conference on Natural
  Language Processing}, pp.\  147--155, 2011.

\bibitem[Narayan et~al.(2018)Narayan, Cohen, and Lapata]{narayan2018don}
Shashi Narayan, Shay~B Cohen, and Mirella Lapata.
\newblock Don't give me the details, just the summary! topic-aware
  convolutional neural networks for extreme summarization.
\newblock \emph{arXiv preprint arXiv:1808.08745}, 2018.

\bibitem[Ng et~al.(2013)Ng, Wu, Wu, Hadiwinoto, and
  Tetreault]{ng-etal-2013-conll}
Hwee~Tou Ng, Siew~Mei Wu, Yuanbin Wu, Christian Hadiwinoto, and Joel Tetreault.
\newblock The {C}o{NLL}-2013 shared task on grammatical error correction.
\newblock In \emph{Proceedings of the Seventeenth Conference on Computational
  Natural Language Learning: Shared Task}, pp.\  1--12, Sofia, Bulgaria, August
  2013. Association for Computational Linguistics.
\newblock URL \url{https://aclanthology.org/W13-3601}.

\bibitem[Ng et~al.(2014)Ng, Wu, Briscoe, Hadiwinoto, Susanto, and
  Bryant]{ng2014conll}
Hwee~Tou Ng, Siew~Mei Wu, Ted Briscoe, Christian Hadiwinoto, Raymond~Hendy
  Susanto, and Christopher Bryant.
\newblock The conll-2014 shared task on grammatical error correction.
\newblock In \emph{Proceedings of the Eighteenth Conference on Computational
  Natural Language Learning: Shared Task}, pp.\  1--14, 2014.

\bibitem[Ott et~al.(2018)Ott, Edunov, Grangier, and Auli]{ott2018scaling}
Myle Ott, Sergey Edunov, David Grangier, and Michael Auli.
\newblock Scaling neural machine translation.
\newblock \emph{arXiv preprint arXiv:1806.00187}, 2018.

\bibitem[Panahi et~al.(2021)Panahi, Saeedi, and Arodz]{panahi2021shapeshifter}
Aliakbar Panahi, Seyran Saeedi, and Tom Arodz.
\newblock Shapeshifter: a parameter-efficient transformer using factorized
  reshaped matrices.
\newblock \emph{Advances in Neural Information Processing Systems}, 34, 2021.

\bibitem[Papineni et~al.(2002)Papineni, Roukos, Ward, and
  Zhu]{papineni2002bleu}
Kishore Papineni, Salim Roukos, Todd Ward, and Wei-Jing Zhu.
\newblock Bleu: a method for automatic evaluation of machine translation.
\newblock In \emph{Proceedings of the 40th annual meeting of the Association
  for Computational Linguistics}, pp.\  311--318, 2002.

\bibitem[Post(2018)]{post2018call}
Matt Post.
\newblock A call for clarity in reporting bleu scores.
\newblock \emph{arXiv preprint arXiv:1804.08771}, 2018.

\bibitem[Raffel et~al.(2019)Raffel, Shazeer, Roberts, Lee, Narang, Matena,
  Zhou, Li, and Liu]{raffel2019exploring}
Colin Raffel, Noam Shazeer, Adam Roberts, Katherine Lee, Sharan Narang, Michael
  Matena, Yanqi Zhou, Wei Li, and Peter~J Liu.
\newblock Exploring the limits of transfer learning with a unified text-to-text
  transformer.
\newblock \emph{arXiv preprint arXiv:1910.10683}, 2019.

\bibitem[Reid et~al.(2021)Reid, Marrese-Taylor, and Matsuo]{reid2021subformer}
Machel Reid, Edison Marrese-Taylor, and Yutaka Matsuo.
\newblock Subformer: Exploring weight sharing for parameter efficiency in
  generative transformers.
\newblock \emph{arXiv preprint arXiv:2101.00234}, 2021.

\bibitem[Sun et~al.(2021)Sun, Ge, Wei, and Wang]{sun2021instantaneous}
Xin Sun, Tao Ge, Furu Wei, and Houfeng Wang.
\newblock Instantaneous grammatical error correction with shallow aggressive
  decoding.
\newblock In \emph{Proceedings of the 59th Annual Meeting of the Association
  for Computational Linguistics and the 11th International Joint Conference on
  Natural Language Processing (Volume 1: Long Papers)}, pp.\  5937--5947, 2021.

\bibitem[Takase \& Kiyono(2021)Takase and Kiyono]{takase2021lessons}
Sho Takase and Shun Kiyono.
\newblock Lessons on parameter sharing across layers in transformers.
\newblock \emph{arXiv preprint arXiv:2104.06022}, 2021.

\bibitem[Tambe et~al.(2021)Tambe, Hooper, Pentecost, Jia, Yang, Donato, Sanh,
  Whatmough, Rush, Brooks, et~al.]{tambe2021edgebert}
Thierry Tambe, Coleman Hooper, Lillian Pentecost, Tianyu Jia, En-Yu Yang, Marco
  Donato, Victor Sanh, Paul Whatmough, Alexander~M Rush, David Brooks, et~al.
\newblock Edgebert: Sentence-level energy optimizations for latency-aware
  multi-task nlp inference.
\newblock In \emph{MICRO-54: 54th Annual IEEE/ACM International Symposium on
  Microarchitecture}, pp.\  830--844, 2021.

\bibitem[Vaswani et~al.(2017)Vaswani, Shazeer, Parmar, Uszkoreit, Jones, Gomez,
  Kaiser, and Polosukhin]{vaswani2017attention}
Ashish Vaswani, Noam Shazeer, Niki Parmar, Jakob Uszkoreit, Llion Jones,
  Aidan~N Gomez, {\L}ukasz Kaiser, and Illia Polosukhin.
\newblock Attention is all you need.
\newblock In \emph{Advances in neural information processing systems}, pp.\
  5998--6008, 2017.

\bibitem[Wang et~al.(2020)Wang, Wu, Liu, Cai, Zhu, Gan, and Han]{wang2020hat}
Hanrui Wang, Zhanghao Wu, Zhijian Liu, Han Cai, Ligeng Zhu, Chuang Gan, and
  Song Han.
\newblock Hat: Hardware-aware transformers for efficient natural language
  processing.
\newblock \emph{arXiv preprint arXiv:2005.14187}, 2020.

\bibitem[Wu et~al.(2019)Wu, Fan, Baevski, Dauphin, and Auli]{wu2018pay}
Felix Wu, Angela Fan, Alexei Baevski, Yann Dauphin, and Michael Auli.
\newblock Pay less attention with lightweight and dynamic convolutions.
\newblock In \emph{International Conference on Learning Representations}, 2019.
\newblock URL \url{https://arxiv.org/abs/1901.10430}.

\bibitem[Wu et~al.(2020)Wu, Liu, Lin, Lin, and Han]{wu2020lite}
Zhanghao Wu, Zhijian Liu, Ji~Lin, Yujun Lin, and Song Han.
\newblock Lite transformer with long-short range attention.
\newblock \emph{arXiv preprint arXiv:2004.11886}, 2020.

\bibitem[Xu et~al.(2020)Xu, Zhou, Ge, Wei, and Zhou]{xu2020bert}
Canwen Xu, Wangchunshu Zhou, Tao Ge, Furu Wei, and Ming Zhou.
\newblock Bert-of-theseus: Compressing bert by progressive module replacing.
\newblock In \emph{Proceedings of the 2020 Conference on Empirical Methods in
  Natural Language Processing (EMNLP)}, pp.\  7859--7869, 2020.

\bibitem[Yannakoudakis et~al.(2011)Yannakoudakis, Briscoe, and
  Medlock]{yannakoudakis2011new}
Helen Yannakoudakis, Ted Briscoe, and Ben Medlock.
\newblock A new dataset and method for automatically grading esol texts.
\newblock In \emph{Proceedings of the 49th annual meeting of the association
  for computational linguistics: human language technologies}, pp.\  180--189,
  2011.

\bibitem[Zhang et~al.(2021)Zhang, Tay, Zhang, Chan, Luu, Hui, and
  Fu]{zhang2021beyond}
Aston Zhang, Yi~Tay, Shuai Zhang, Alvin Chan, Anh~Tuan Luu, Siu~Cheung Hui, and
  Jie Fu.
\newblock Beyond fully-connected layers with quaternions: Parameterization of
  hypercomplex multiplications with $1/n $ parameters.
\newblock \emph{arXiv preprint arXiv:2102.08597}, 2021.

\bibitem[Zhou et~al.(2021)Zhou, Ge, Xu, Xu, and Wei]{zhou2021improving}
Wangchunshu Zhou, Tao Ge, Canwen Xu, Ke~Xu, and Furu Wei.
\newblock Improving sequence-to-sequence pre-training via sequence span
  rewriting.
\newblock In \emph{Proceedings of the 2021 Conference on Empirical Methods in
  Natural Language Processing}, pp.\  571--582, 2021.

\end{thebibliography}
\bibliographystyle{iclr2022_conference}

\clearpage
\appendix

\section{Details of Evaluations for MT and GEC}\label{sec:app_train}

For MT, we follow the setting of \citet{ott2018scaling} to train the model on the WMT14 datasets. For En-De, the training set contains 4.5M parallel sentence pairs. We use newstest2013 as our dev set. For En-Fr, there are 36M parallel sentence pairs for training, and we use newstest2012+2013 as the dev set.

In GEC evaluation, we follow previous work~\cite{chen-etal-2020-improving-efficiency,zhou2021improving} to use the BEA-19 restricted setting, training with Lang-8~\citep{mizumoto2011mining}, FCE~\citep{yannakoudakis2011new} and WI+LOCNESS~\citep{bryant2019bea}. We validate on CoNLL-13 shared task dataset~\citep{ng-etal-2013-conll}, and test on CoNLL-14 shared task dataset~\citep{ng2014conll}. After de-duplicating, we have around 900K sentence pairs for training. Both the dev (CoNLL-13) and test (CoNLL-14) have 1.3K sampales. We train a sentencepiece~\citep{kudo2018sentencepiece} model of 2K vocabulary for tokenization, and evaluate with the metric of Max-Match~\citep{dahlmeier2012better} $F_{0.5}$. 

For model training configuration, we show in Table \ref{tab:train}; The ablation study into the reduction of $d_{\textrm{decffn}}$ is presented in Table \ref{tab:decffn}.

\begin{table}[h]
\centering
\small
\scalebox{0.86}{
\begin{tabular} {lr} 
\hline
Configurations	         &	Values
\\
\hline 
Number of epochs		& 1000				\\
Devices                 & 8 Nvidia V100 GPU      \\
Max tokens per GPU		& 20,000					\\
Update Frequency        & 4                     \\
Optimizer 				& Adam 					\\
						& ($\beta_1$=0.9, $\beta_2$=0.99, $\epsilon$=$1\times10^{-8}$)	\\
Learning rate 			& $1\times10^{-3}$ \\
Learning rate scheduler & inverse sqrt \\ 
Warmup                  & 4000 \\
Weight decay            & 0.00001 \\
Loss Function 			& label smoothed cross entropy \\
						& (label-smoothing=0.1) \\
Dropout 				& [0.1, 0.2] for MT, [0.3, 0.4, 0.5] for GEC \\
\hline
\end{tabular}
}
\caption{Training details for \textsc{EdgeFormer} for NMT and GEC.}\label{tab:train}
\end{table}

\begin{table}[h]
\centering
\begin{tabular}{c|c|c}
\toprule
$d_{\textrm{decffn}}$ & \#Param & sacreBLEU  \\ \midrule
2048 & 46M & 26.5     \\
512 & 37M & 26.4     \\
256 & 35M & 26.4     \\
128 & 34M & 26.4     \\ \bottomrule
\end{tabular}
\caption{The ablation study into the reduction of $d_{\textrm{decffn}}$ on a standard 6+6 Transformer on the dev set.}\label{tab:decffn}
\end{table}

\begin{table}[t]
\centering
\small
\begin{tabular} {lr} 
\hline
Configurations	         &	Values
\\
\hline 
Total updates		& 250,000				\\
Devices                 & $8 \times 8$ Nvidia V100 GPU      \\
Batch size per GPU		& 128					\\
Sample length		& 512				\\
Optimizer 				& Adam 					\\
						& ($\beta_1$=0.9, $\beta_2$=0.98 $\epsilon$=$1\times10^{-6}$)	\\
Learning rate 			& $5\times10^{-4}$ \\
Learning rate scheduler & polynomial \\ 
clip norm   & 2.0 \\
Warmup                  & 10,000 \\
Loss Function 			& cross entropy \\
Weight decay            & 0.0 \\
Dropout 				& 0.1 \\
\hline
\end{tabular}
\caption{Pretraining details for \textsc{EdgeLM}.}\label{tab:pretrain_config}
\end{table}

\begin{table}[t]
\centering
\small
\scalebox{0.92}{
\begin{tabular} {lr} 
\hline
Configurations	         &	Values
\\
\hline 
Total updates		& 100,000				\\
Devices                 & $8$ Nvidia V100 GPU      \\
Max tokens per GPU		& 20,000					\\
Optimizer 				& Adam 					\\
						& ($\beta_1$=0.9, $\beta_2$=0.98 $\epsilon$=$1\times10^{-6}$)	\\
Learning rate 			& Vary for different downstream tasks \\
Learning rate scheduler & polynomial \\ 
clip norm   & 1.0 \\
Warmup                  & 8,000 \\
Loss Function 			& cross entropy \\
Weight decay            & 0.0 \\
Dropout 				& 0.1 \\
\hline
\end{tabular}}
\caption{Fine-tuning details for \textsc{EdgeLM}.}\label{tab:ft_config}
\end{table}

\section{Configurations of Pretraining and Fine-tuning}\label{sec:app_pretrain}
We pretrain \textsc{EdgeFormer} with the same pretrain data as RoBERTa~\citep{liu2019roberta},  through the same pretrain task as T5~\citep{raffel2019exploring}. The detailed configuration of pretraining is shown in Table \ref{tab:pretrain_config}.

For downstream task fine-tuning, we present the configuration details in Table \ref{tab:ft_config}.

\begin{table*}[t]
\centering
\small
\scalebox{0.95}{
\begin{tabular}{l|c|c|c|ccc|ccc}
\toprule
\multirow{2}{*}{Model}           & \multirow{2}{*}{\#Param} & \multirow{2}{*}{FLOPS} & CoNLL-14  & \multicolumn{3}{c|}{Xsum} & \multicolumn{3}{c}{QG} \\
                                 &                          &                        & $F_{0.5}$ & RG-1   & RG-2   & RG-L   & B4     & MTR   & RG-L  \\ \midrule
Transformer                      & 44M                      & 1.8G                   & 50.1 & 31.2   & 10.7   & 24.9   & 2.6    & 9.0   & 26.0  \\ \midrule
Pretrained UT ($d=512$, $d_{\textrm{ffn}}=3072$)            & 9.4M                     & 1.9G                   & 51.1 & 36.7   & 14.9   & 29.7   & 20.1   & 22.4  & 47.1  \\
\textsc{EdgeLM} ($d=512$)    & 9.4M                     & 1.3G                   & \bf 52.0 & \bf 37.2   & \bf 15.4   & \bf 30.3   & \bf 20.6   & \bf 23.1  & \bf 47.4  \\ \midrule
Pretrained UT ($d=384$, $d_{\textrm{ffn}}=2432$) & 5.5M                     & 1.1G                   & 50.1 & 31.9   & 11.5   & 25.7   & 17.9   & 21.0  & 45.9  \\
\textsc{EdgeLM} ($d=384$)    & 5.5M                     & 0.8G                   & 50.1 & \bf 32.3   & \bf 11.7   & \bf 26.0   & \bf 18.9   & \bf 21.2  & \bf 45.9  \\ \midrule
Pretrained UT ($d=768$, $d_{\textrm{ffn}}=4608$) & 21.2M                    & 4.2G                   & 52.8 & 37.3   & 15.6   & 30.4   & 20.9   & 23.5  & 47.5  \\
\textsc{EdgeLM} ($d=768$)    & 21.4M                    & 2.9G                   & \bf 53.1 & \bf 37.9   & \bf 15.9   & \bf 30.8   & \bf 21.0   & \bf 23.6  & \bf 47.8 \\ \bottomrule
\end{tabular}
}
\caption{The comparison between pretrained Universal Transformer (UT) and EdgeLM with Adapter-based layer adaptation ($r=d/8$). We enlarge UT's $d_{\textrm{ffn}}$ to let its model size comparable with its EdgeFormer counterpart with the same $d$ for fair comparison given the same parameter budget though introducing additional cost.}\label{tab:pretrain_size}
\end{table*}

\begin{table*}[t]
\centering
\small
\begin{tabular}{c|c|c|c|ccc|ccc}
\toprule
\multirow{2}{*}{\textbf{Vocab}} & \multirow{2}{*}{$d_{\textrm{embed}}$} & \multirow{2}{*}{\textbf{\#Param (including embedding)}} & \textbf{CoNLL-14} & \multicolumn{3}{c|}{\textbf{Xsum}} & \multicolumn{3}{c}{\textbf{QG}} \\
                                &                                  &                                   & $F_{0.5}$             & RG-1      & RG-2      & RG-L      & B4        & MTR      & RG-L     \\ \midrule
spm2k                           & 512                              & 10.4M                              & \bf 52.7              & 36.3      & 14.8      & 29.5      & 19.0      & 21.7     & 46.3     \\
spm8k                           & 128                              & 10.5M                              & 52.0              & \bf 37.2      & \bf 15.4      & \bf 30.3      & \bf 20.6      & \bf 23.1     & \bf 47.4     \\
spm16k                          & 64                               & 10.5M                              & 51.6              & 37.1      & 15.1      & 30.0      & 19.6      & 22.3     & 46.7    \\ \bottomrule
\end{tabular}
\caption{\textsc{EdgeLM} ($d=512$) with Adapter-LA ($r=32$) using different vocabulary and $d_{\textrm{embed}}$. Except the model with spm2k whose $d_{\textrm{embed}}=d$, the models with spm8k and spm16k use factorized embedding to prevent increasing the total parameters.}\label{tab:pretrain_embed}
\end{table*}

\section{Detailed Evaluation of EdgeLM}
In addition to the evaluation results presented in Table \ref{tab:pretrain}, we present more detailed results in Table \ref{tab:pretrain_size} and \ref{tab:pretrain_embed} to show the performance of \textsc{EdgeLM} in different sizes and with different vocabularies with factorized embedding parameterization respectively. According to Table \ref{tab:pretrain_size}, \textsc{EdgeLM}s consistently outperform the pretrained UTs in the downstream tasks under various model sizes with only 70\% computation cost of UTs.

According to Table \ref{tab:pretrain_embed}, we show that enlarging the vocabulary size with factorized embedding is effective to improve abstractive summarization and question generation tasks, while it appears to have an adverse effect for GEC. One reason for the performance degradation is that GEC is more sensitive to morphological and syntactic information of a token; when $d_\textrm{embed}$ becomes small with factorized embedding, it may not accurately capture the morphological and syntactic information of the token.

\end{document}